\documentclass[11pt,a4paper]{article}
\usepackage{acl2019}
\usepackage{times}
\usepackage{latexsym}

\usepackage{url}

\usepackage{bm}
\usepackage{amsmath}
\usepackage{amssymb}
\usepackage{multirow}
\usepackage{graphicx}
\usepackage{subfigure}
\usepackage[T1]{fontenc}
\usepackage[utf8]{inputenc}
\usepackage{CJKutf8}

\def\argmax{\mathop{\rm argmax}}%

\newcommand{\tabincell}[2]{\begin{tabular}{@{}#1@{}}#2\end{tabular}}

\def\vh{\bm{h}}
\def\vv{\bm{v}}
\def\vw{\bm{x}}

\def\vz{\bm{z}}
\def\vx{\bm{x}}

\aclfinalcopy 

\title{Incorporating Sememes into Chinese Definition Modeling}

\author{Liner Yang$^\dagger$, Cunliang Kong$^\dagger$, Yun Chen$^\ddagger$, Yang Liu$^\S$,  Qinan Fan$^\dagger$, Erhong Yang$^\dagger$\\
	$^\dagger$Beijing Language and Culture University, Beijing, China \\
	$^\ddagger$The University of Hong Kong, Hong Kong, China \\
	$^\S$Tsinghua University, Beijing, China \\
    }

\date{}

\begin{document}
\maketitle
\begin{abstract}
  Chinese definition modeling is a challenging task that generates a dictionary definition in Chinese for a given Chinese word. To accomplish this task, we construct the Chinese Definition Modeling Corpus (CDM), which contains triples of word, sememes and the corresponding definition. We present two novel models to improve Chinese definition modeling: the Adaptive-Attention model (AAM) and the Self- and Adaptive-Attention Model (SAAM). AAM successfully incorporates sememes for generating the definition with an adaptive attention mechanism. It has the capability to decide which sememes to focus on and when to pay attention to sememes. SAAM further replaces recurrent connections in AAM with self-attention and relies entirely on the attention mechanism, reducing the path length between word, sememes and definition. Experiments on CDM demonstrate that by incorporating sememes, our best proposed model can outperform the state-of-the-art method by +6.0 BLEU.
\end{abstract}

\begin{CJK*}{UTF8}{gbsn}
\section{Introduction}
Chinese definition modeling is the task of generating a definition in Chinese for a given Chinese word. This task can benefit the compilation of dictionaries, especially dictionaries for Chinese as a foreign language (CFL) learners.

In recent years, the number of CFL learners has risen sharply. 
In 2017, 770,000 people took the Chinese Proficiency Test, an increase of 38\% from 2016\footnote{According to Hanban, \url{http://www.hanban.edu.cn/tests/node_7475.htm}}. However, most Chinese dictionaries are for native speakers. 
Since these dictionaries usually require a fairly high level of Chinese, it is necessary to build a dictionary specifically for CFL learners. Manually writing definitions relies on the knowledge of lexicographers and linguists, which is expensive and time-consuming \cite{Wang:2009, Zhang:2011, jin2014review}. 
Therefore, the study on writing definitions automatically is of practical significance.

Definition modeling was first proposed by \citet{Noraset+:2017} as a tool to evaluate different word embeddings. \citet{Gadetsky+:2018} extended the work by incorporating word sense disambiguation to generate context-aware word definition. Both methods are based on recurrent neural network encoder-decoder framework without attention. In contrast, this paper formulates the definition modeling task as an automatic way to accelerate dictionary compilation.

In this work, we introduce a new dataset for the Chinese definition modeling task that we call Chinese Definition Modeling Corpus \hyperlink{cdm}{(CDM)}. CDM consists of 104,517 entries, where each entry contains a word, the sememes of a specific word sense, and the definition in Chinese of the same word sense. Sememes are \textbf{minimum semantic units} of word meanings, and the meaning of each word sense is typically composed of several sememes, as is illustrated in Figure \ref{fig:example}. For a given word sense, CDM annotates the sememes according to HowNet \cite{zhendong2006hownet}, and the definition according to Chinese Concept Dictionary (CCD) \cite{yang+yu:2017}. Since sememes have been widely used in improving word representation learning \cite{Niu+:2017} and word similarity computation \cite{Liu+Li:2002}, we argue that sememes can benefit the task of definition modeling. 

We propose two novel models to incorporate sememes into Chinese definition modeling: the Adaptive-Attention Model (AAM) and the Self- and Adaptive-Attention Model (SAAM). Both models are based on the encoder-decoder framework. The encoder maps word and sememes into a sequence of continuous representations, and the decoder then attends to the output of the encoder and generates the definition one word at a time. Different from the vanilla attention mechanism, the decoder of both models employs the \textit{adaptive attention mechanism} to decide which sememes to focus on and when to pay attention to sememes at one time \cite{Lu+:2017}. Following \citet{Noraset+:2017,Gadetsky+:2018}, the AAM is built using recurrent neural networks (RNNs). However, recent works demonstrate that attention-based architecture that entirely eliminates recurrent connections can obtain new state-of-the-art in neural machine translation \cite{Vaswani+:2017}, constituency parsing \cite{Kitaev+klein:2018} and semantic role labeling \cite{Tan+:2018}. 
In the SAAM, we replace the LSTM-based encoder and decoder with an architecture based on self-attention. This fully attention-based model allows for more parallelization, reduces the path length between word, sememes and the definition, and can reach a new state-of-the-art on the definition modeling task. 
To the best of our knowledge, this is the first work to introduce the attention mechanism and utilize external resource for the definition modeling task.

In experiments on the CDM dataset we show that our proposed AAM and SAAM outperform the state-of-the-art approach with a large margin. By efficiently incorporating sememes, the SAAM achieves the best performance with improvement over the state-of-the-art method by +6.0 BLEU.

\section{Methodology}
The definition modeling task is to generate an explanatory sentence for the interpreted word.
For example, given the word ``旅馆'' (hotel), 
a model should generate a sentence like this: ``给旅行者提供食宿和其他服务的地方'' (A place to provide residence and other services for tourists). 
Since distributed representations of words have been shown to capture lexical syntax and semantics, 
it is intuitive to employ word embeddings to generate natural language definitions. 

Previously, \citet{Noraset+:2017} proposed several model architectures to generate a definition according to the distributed representation of a word. 
We briefly summarize their model with the best performance in Section \ref{sec:base} and adopt it as our baseline model.

\begin{figure}[tb]
    \centering
    \includegraphics[width=0.47\textwidth]{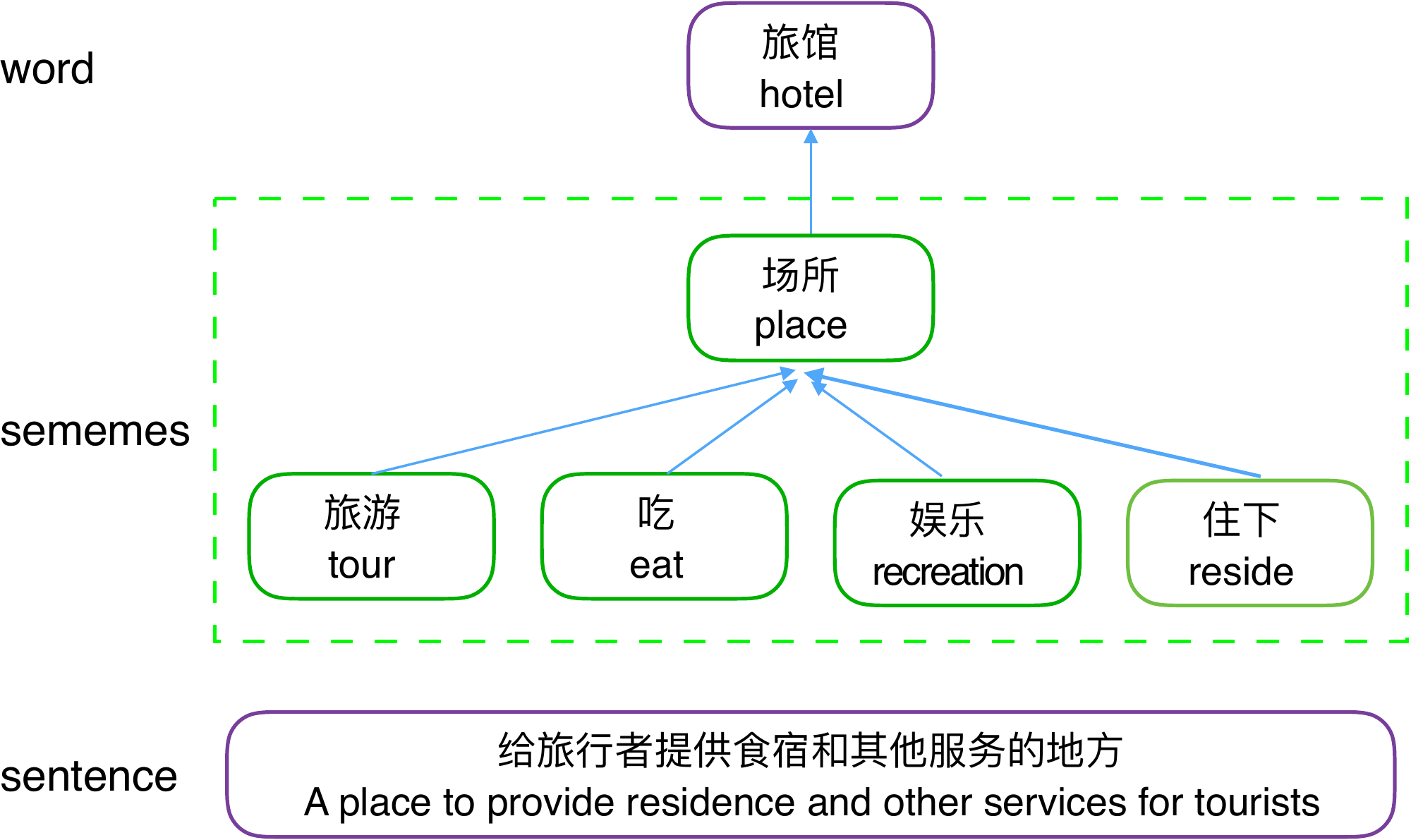}
    \caption{An example of the CDM dataset. The word ``旅馆'' (hotel) has five sememes, which are ``场所'' (place), ``旅游'' (tour), ``吃'' (eat), ``娱乐'' (recreation) and ``住下'' (reside).}\label{fig:example}
\end{figure}

Inspired by the works that use sememes to improve word representation learning \cite{Niu+:2017} and word similarity computation \cite{Liu+Li:2002}, 
we propose the idea of incorporating sememes into definition modeling. 
Sememes can provide additional semantic information for the task. 
As shown in Figure \ref{fig:example}, sememes are highly correlated to the definition. 
For example, the sememe ``场所'' (place) is related with the word ``地方'' (place) of the definition, and the sememe ``旅游'' (tour) is correlated to the word ``旅行者'' (tourists) of the definition.

Therefore, to make full use of the sememes in CDM dataset, we propose AAM and SAAM for the task, in Section \ref{sec:aam} and Section \ref{sec:saam}, respectively.

\subsection{Baseline Model} \label{sec:base}
The baseline model \cite{Noraset+:2017} is implemented with a recurrent neural network based encoder-decoder framework. Without utilizing the information of sememes, it learns a probabilistic mapping $P(y | x)$ from the word $x$ to be defined to a definition $y = \lbrack y_1, \dots, y_T \rbrack$, in which $y_t$ is the $t$-th word of definition $y$.

More concretely, given a word $x$ to be defined, the encoder reads the word and generates its word embedding $\vx$ as the encoded information. 
Afterward, the decoder computes the conditional probability of each definition word $y_t$ depending on the previous definition words $y_{<t}$, as well as the word being defined $x$, i.e., $P(y_t|y_{<t},x)$. $P(y_t|y_{<t},x)$ is given as:
\begin{eqnarray}
& P(y_t|y_{<t},x) \propto \exp{(y_t;\bm{z}_t,\bm{x})} & \\
& \bm{z}_t = f(\bm{z}_{t-1},y_{t-1},\bm{x}) &
\end{eqnarray}
where $\bm{z}_t$ is the decoder's hidden state at time $t$, $f$ is a recurrent nonlinear function such as LSTM and GRU, and $\bm{x}$ is the embedding of the word being defined. Then the probability of $P(y | x)$ can be computed according to the probability chain rule:
\begin{eqnarray}
P(y | x) = \prod_{t=1}^{T} P(y_t|y_{<t},x) 
\end{eqnarray}
We denote all the parameters in the model as $\theta$ and the definition corpus as $D_{x,y}$, which is a set of word-definition pairs. 
Then the model parameters can be learned by maximizing the log-likelihood:
\begin{eqnarray}
\hat{\theta} = \argmax_{\theta} \sum_{\langle x, y \rangle \in D_{x,y}}\log P(y | x; \theta) \label{eq:mle_base}
\end{eqnarray}

\subsection{Adaptive-Attention Model} \label{sec:aam}
Our proposed model aims to incorporate sememes into the definition modeling task. Given the word to be defined $x$ and its corresponding sememes $s=\lbrack s_1, \dots, s_N \rbrack$, we define the probability of generating the definition $y=\lbrack y_1, \dots, y_t \rbrack$ as:
\begin{eqnarray}
P(y | x, s) = \prod_{t=1}^{T} P(y_t|y_{<t},x,s) \label{eqn:aam_p}
\end{eqnarray}
Similar to Eq. \ref{eq:mle_base}, we can maximize the log-likelihood with the definition corpus $D_{x,s,y}$ to learn model parameters:
\begin{eqnarray}
\hat{\theta} = \argmax_{\theta} \sum_{\langle x,s,y \rangle \in D_{x,s,y}}\log P(y | x, s; \theta) \label{eq:mle_aam}
\end{eqnarray}
The probability $P(y | x, s)$ can be implemented with an adaptive attention based encoder-decoder framework, which we call Adaptive-Attention Model (AAM). The new architecture consists of a bidirectional RNN as the encoder and a RNN decoder that adaptively attends to the sememes during decoding a definition.  

\paragraph{Encoder}
Similar to \citet{Bahdanau2015NeuralMT}, the encoder is a bidirectional RNN, consisting of forward and backward RNNs. Given the word to be defined $x$ and its corresponding sememes $s=\lbrack s_1, \dots, s_N \rbrack$, we define the input sequence of vectors for the encoder as $\bm{v}=[\bm{v}_1,\dots,\bm{v}_{N}]$. The vector $\bm{v}_n$ is given as follows:
\begin{eqnarray}
\bm{v}_n = \lbrack \bm{x}; \bm{s}_n \rbrack
\end{eqnarray}
where $\bm{x}$ is the vector representation of the word $x$, $\bm{s}_n$ is the vector representation of the $n$-th sememe $s_n$,  and $\lbrack \bm{a};\bm{b}\rbrack $ denote concatenation of vector $\bm{a}$ and $\bm{b}$.

The forward RNN $\overrightarrow{f}$ reads the input sequence of vectors from $\bm{v}_1$ to $\bm{v}_N$ and calculates a forward hidden state for position $n$ as:
\begin{eqnarray}
\overrightarrow{\vh_{n}} &=& f(\vv_n, \overrightarrow{\vh_{n-1}})
\end{eqnarray}
where $f$ is an LSTM or GRU. Similarly, the backward RNN $\overleftarrow{f}$ reads the input sequence of vectors from $\bm{v}_N$ to $\bm{v}_1$ and obtain a backward hidden state for position $n$ as:
\begin{eqnarray}
\overleftarrow{\vh_{n}} &=& f(\vh_n, \overleftarrow{\vh_{n+1}})
\end{eqnarray}
In this way, we obtain a sequence of encoder hidden states $\vh=\left[\vh_1,...,\vh_N\right]$, by concatenating the forward hidden state $\overrightarrow{\vh_{n}}$ and the backward one $\overleftarrow{\vh_{n}}$ at each position $n$:
\begin{eqnarray}
\vh_n=\left[\overrightarrow{\vh_{n}}, \overleftarrow{\vh_{n}}\right]
\end{eqnarray}    
The hidden state $\vh_n$ captures the sememe- and word-aware information of the $n$-th sememe.

\begin{figure*}[!tb]
    \centering
    \includegraphics[width=0.7\textwidth]{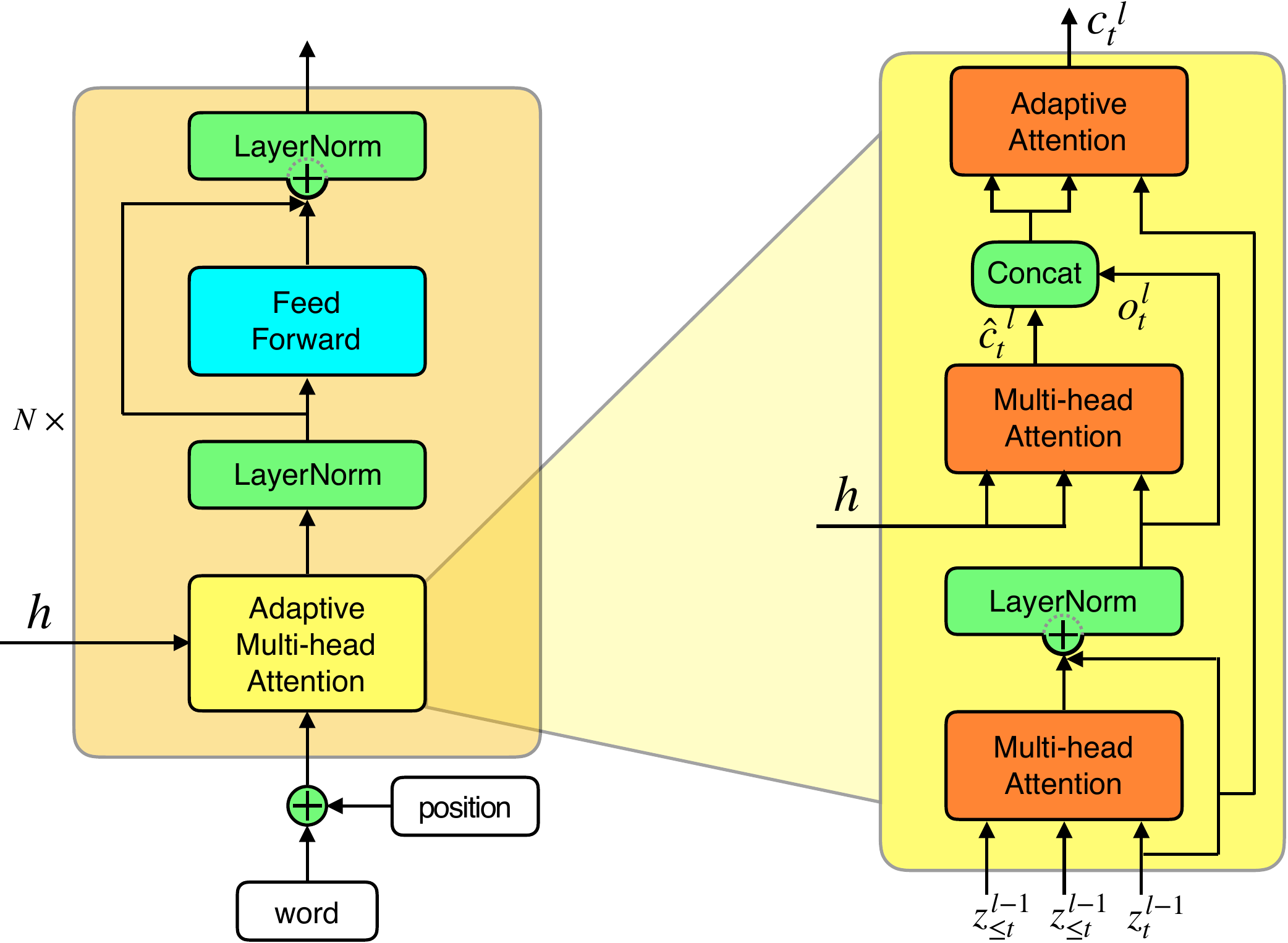}
    \caption{An overview of the decoder for the SAAM. The left sub-figure shows our decoder contains $N$ identical layers, where each layer contains two sublayer: adaptive multi-head attention layer and feed-forward layer. The right sub-figure shows how we perform the adaptive multi-head attention at layer $l$ and time $t$ for the decoder. $\bm{z}_t^l$ represents the hidden state of the decoder at layer $l$, time $t$. $\bm{h}$ denotes the output from the encoder stack. $\hat{\bm{c}_t}^l$ is the sememe context, while $\bm{o}_t^l$ is the LM context. $\bm{c}_t^l$ is the output of the adaptive multi-head attention layer at time $t$.}\label{fig:architecture}
    \vspace{-0.4cm}
\end{figure*}

\paragraph{Decoder}
As attention-based neural encoder-decoder frameworks have shown great success in image captioning \cite{Xu2015ShowAA}, document summarization \cite{See2017GetTT} and neural machine translation \cite{Bahdanau2015NeuralMT}, it is natural to adopt the attention-based recurrent decoder in \citet{Bahdanau2015NeuralMT} as our decoder. 
The vanilla attention attends to the sememes at every time step. 
However, not all words in the definition have corresponding sememes. 
For example, sememe {``住下'' (reside)} could be useful when generating {``食宿'' (residence)}, but none of the sememes is useful when generating {``提供'' (provide)}. 
Besides, language correlations make the sememes unnecessary when generating words like {``和'' (and)} and {``给'' (for)}.

Inspired by \citet{Lu+:2017}, we introduce the adaptive attention mechanism for the decoder. At each time step $t$, we summarize the time-varying sememes' information as sememe context, and the language model's information as LM context. Then, we use another attention to obtain the context vector, relying on either the sememe context or LM context. 

More concretely, we define each conditional probability in Eq. \ref{eqn:aam_p} as:
\begin{eqnarray}
& P(y_t|y_{<t},x,s) \propto \exp{(y_t;\bm{z}_t,\bm{c}_t)} & \label{eq:aam_out} \\
& \bm{z}_t = f(\bm{z}_{t-1},y_{t-1},\bm{c}_t) & \label{eq:aam_upz}
\end{eqnarray}
where $\bm{c}_t$ is the context vector from the output of the adaptive attention module at time $t$, $\bm{z}_t$ is a decoder's hidden state at time $t$.

To obtain the context vector $\bm{c}_t$, we first compute the sememe context vector $\hat{\bm{c}_t}$ and the LM context $\bm{o}_t$. Similar to the vanilla attention, the sememe context $\hat{\bm{c}_t}$ is obtained with a soft attention mechanism as:
\begin{eqnarray}
\hat{\bm{c}_t} = \sum_{n=1}^{N} \alpha_{tn} \vh_n,
\end{eqnarray}
where
\begin{eqnarray}
\alpha_{tn} &=& \frac{\mathrm{exp}(e_{tn})}{\sum_{i=1}^{N} \mathrm{exp}(e_{ti})} \nonumber \\
e_{tn} &=& \bm{w}_{\hat{c}}^T[\vh_n; \vz_{t-1}].
\end{eqnarray}
Since the decoder's hidden states store syntax and semantic information for language modeling, we compute the LM context $\bm{o}_t$ with a gated unit, whose input is the definition word $y_t$ and the previous hidden state $\vz_{t-1}$:
\begin{eqnarray}
\bm{g}_t &=& \sigma (\bm{W}_g [y_{t-1}; \vz_{t-1}] + \bm{b}_g) \nonumber \\
\bm{o}_t &=& \bm{g}_t \odot \mathrm{tanh} (\vz_{t-1}) \end{eqnarray}
Once the sememe context vector $\hat{\bm{c}_t}$ and the LM context $\bm{o}_t$ are ready, we can generate the context vector with an adaptive attention layer as:
\begin{eqnarray}
\bm{c}_t = \beta_t \bm{o}_t + (1-\beta_t)\hat{\bm{c}_t}, \label{eq:context_q1}
\end{eqnarray}
where 
\begin{eqnarray}
\beta_{t} &=& \frac{\mathrm{exp}(e_{to})}{\mathrm{exp}(e_{to})+\mathrm{exp}(e_{t\hat{c}})} \nonumber \\
e_{to} &=& (\bm{w}_c)^T[\bm{o}_t;\bm{z}_t] \nonumber \\
e_{t\hat{c}} &=& (\bm{w}_c)^T[\hat{\bm{c}_t};\bm{z}_t] \label{eq:context_q2}
\end{eqnarray}
$\beta_{t}$ is a scalar in range $[0,1]$, which controls the relative importance of LM context and sememe context.

Once we obtain the context vector $\bm{c}_t$, we can update the decoder's hidden state and generate the next word according to Eq. \ref{eq:aam_upz} and Eq. \ref{eq:aam_out}, respectively. 

\subsection{Self- and Adaptive-Attention Model} \label{sec:saam}

Recent works demonstrate that an architecture entirely based on attention can obtain new state-of-the-art in neural machine translation \cite{Vaswani+:2017}, constituency parsing \cite{Kitaev+klein:2018} and semantic role labeling \cite{Tan+:2018}. SAAM adopts similar architecture and replaces the recurrent connections in AAM with self-attention. Such architecture not only reduces the training time by allowing for more parallelization, but also learns better the dependency between word, sememes and tokens of the definition by reducing the path length between them. 

\paragraph{Encoder}
Given the word to be defined $x$ and its corresponding ordered sememes $s=[s_1, \dots, s_{N}]$, we combine them as the input sequence of embeddings for the encoder, i.e., $\bm{v}=[\vv_0, \vv_1, \dots, \vv_{N}]$. The $n$-th vector $\vv_n$ is defined as:
\begin{eqnarray}
\vv_n =
\begin{cases}
\vw, &n=0 \cr
\bm{s}_n, &n>0
\end{cases}
\end{eqnarray}
where $\vw$ is the vector representation of the given word $x$, and $\bm{s}_n$ is the vector representation of the $n$-th sememe $s_n$. 

Although the input sequence is not time ordered, position $n$ in the sequence carries some useful information. First, position $0$ corresponds to the word to be defined, while other positions correspond to the sememes. Secondly, sememes are sorted into a logical order in HowNet. 
For example, as the first sememe of the word {``旅馆'' (hotel)}, the sememe {``场所'' (place)} describes its most important aspect, namely, the definition of {``旅馆'' (hotel)} should be {``…… 的地方'' (a place for ...)}.
Therefore, we add learned position embedding to the input embeddings for the encoder:
\begin{eqnarray}
\bm{v}_n = \vv_n + \bm{p}_n
\end{eqnarray}
where $\bm{p}_n$ is the position embedding that can be learned during training.

Then the vectors $\bm{v}=[\vv_0, \vv_1, \dots, \vv_{N}]$ are transformed by a stack of identical layers, where each layers consists of two sublayers: multi-head self-attention layer and position-wise fully connected feed-forward layer. Each of the layers are connected by residual connections, followed by layer normalization \cite{ba2016layer}. We refer the readers to \cite{Vaswani+:2017} for the detail of the layers. The output of the encoder stack is a sequence of hidden states, denoted as $\bm{h}=[\vh_0, \vh_1, \dots, \vh_{N}]$.

\paragraph{Decoder}
The decoder is also composed of a stack of identical layers. In \cite{Vaswani+:2017}, each layer includes three sublayers: masked multi-head self-attention layer, multi-head attention layer that attends over the output of the encoder stack and position-wise fully connected feed-forward layer. In our model, we replace the two multi-head attention layers with an adaptive multi-head attention layer. Similarly to the adaptive attention layer in AAM, the adaptive multi-head attention layer can adaptivelly decide which sememes to focus on and when to attend to sememes at each time and each layer. Figure \ref{fig:architecture} shows the architecture of the decoder.

Different from the adaptive attention layer in AAM that uses single head attention to obtain the sememe context and gate unit to obtain the LM context, the adaptive multi-head attention layer utilizes multi-head attention to obtain both contexts. Multi-head attention performs multiple single head attentions in parallel with linearly projected keys, values and queries, and then combines the outputs of all heads to obtain the final attention result. We omit the detail here and refer the readers to \cite{Vaswani+:2017}. Formally, given the hidden state $\vz_t^{l-1}$ at time $t$, layer $l-1$ of the decoder, we obtain the LM context with multi-head self-attention:
\begin{eqnarray}
\bm{o}_t^l = \textit{MultiHead}(\vz_t^{l-1},\vz_{\leq t}^{l-1},\vz_{\leq t}^{l-1})
\end{eqnarray}  
where the decoder's hidden state $\vz_t^{l-1}$ at time $t$, layer $l-1$ is the query, and $\vz_{\leq t}^{l-1}=[\vz_1^{l-1},...,\vz_t^{l-1}]$, the decoder's hidden states from time $1$ to time $t$ at layer $l-1$, are the keys and values. To obtain better LM context, we employ residual connection and layer normalization after the multi-head self-attention. Similarly, the sememe context can be computed by attending to the encoder's outputs with multi-head attention:
\begin{eqnarray}
\hat{\bm{c}_t}^l = \textit{MultiHead}(\bm{o}_t^l,\bm{h},\bm{h}) 
\end{eqnarray}  
where $\bm{o}_t^l$ is the query, and the output from the encoder stack  $\bm{h}=[\vh_0, \vh_1, \dots, \vh_{N}]$, are the values and keys. 

Once obtaining the sememe context vector $\hat{\bm{c}_t}^l$ and the LM context $\bm{o}_t^l$, we compute the output from the adaptive attention layer with:
\begin{eqnarray}
\bm{c}_t^l = \beta_t^l \bm{o}_t^l + (1-\beta_t^l)\hat{\bm{c}_t}^l, \label{eq:context_q1}
\end{eqnarray}
where 
\begin{eqnarray}
\beta_{t}^l &=& \frac{\mathrm{exp}(e_{to})}{\mathrm{exp}(e_{to})+\mathrm{exp}(e_{t\hat{c}})} \nonumber \\
e_{to}^l &=& (\bm{w}_c^l)^T[\bm{o}_t^l;\bm{z}_t^{l-1}] \nonumber \\
e_{t\hat{c}}^l &=& (\bm{w}_c^l)^T[\hat{\bm{c}_t}^l;\bm{z}_t^{l-1}] \label{eq:context_q2}
\end{eqnarray}

\section{Experiments}
In this section, we will first introduce the construction process of the CDM dataset, then the experimental results and analysis.

\begin{table}[!t]
    \setlength{\tabcolsep}{8pt}
    \centering
    \begin{tabular}{l|c|c|c}
        \hline
        & Train & Valid & Test \\
        \hline 
        \# words & 27,047 & 1,503 & 1,502 \\
        \# entries & 94,029 & 5,218 & 5,270 \\
        \# tokens & 662,410 & 37,174 & 36,813 \\
        \# sememes & 160,792 & 8,966 & 8,851 \\
        \hline 
    \end{tabular}
    \caption{Statistics of the CDM dataset. Jieba Chinese text segmentation tool is used during segmentation. } \label{table:cdds}
\end{table}

\subsection{Dataset}
To verify our proposed models, we construct the CDM dataset for the Chinese definition modeling task. \hypertarget{cdm}{Each entry in the dataset is a triple that consists of: the interpreted word, sememes and a definition for a specific word sense, where the sememes are annotated with HowNet \cite{zhendong2006hownet}, and the definition are annotated with Chinese Concept Dictionary (CCD) \cite{yang+yu:2017}}.

Concretely, for a common word in HowNet and CCD, we first align its definitions from CCD and sememe groups from HowNet, where each group represents one word sense. We define the sememes of a definition as the combined sememes associated with any token of the definition. Then for each definition of a word, we align it with the sememe group that has the largest number of overlapping sememes with the definition's sememes. With such aligned definition and sememe group, we add an entry that consists of the word, the sememes of the aligned sememe group and the aligned definition. Each word can have multiple entries in the dataset, especially the polysemous word. To improve the quality of the created dataset, we filter out entries that the definition contains the interpreted word, or the interpreted word is among function words, numeral words and proper nouns.

After processing, we obtain the dataset that contains 104,517 entries with 30,052 unique interpreted words. We divide the dataset according to the unique interpreted words into training set, validation set and test set with a ratio of 18:1:1. Table \ref{table:cdds} shows the detailed data statistics.
      \vspace{-0.2cm}

\begin{table*}[!t]
    \centering
    \setlength{\tabcolsep}{4pt}
    \begin{tabular}{c|l|l|l}
        \hline
        Word & Sememes  &  Model & \multicolumn{1}{c}{Generated Definitions} \\
        \hline
        \multirow{5}{*}{\tabincell{c}{气压计 \\ (barometer)}}  & \multirow{5}{*}{\tabincell{l}{用具(tool) \\测量(measure) \\力量(strength) \\气(gas)  }} & Baseline & \tabincell{l}{测量轨道刻度盘的仪表 \\ (Instrument for measuring track dial.)} \\
        \cline{3-4}
        & & AAM & \tabincell{l}{测量大气压力的装置 \\ (An instrument that measures atmospheric pressure.)} \\
        \cline{3-4}
        & & SAAM &\tabincell{l}{测量大气压力的装置 \\ (An instrument that measures atmospheric pressure.)}  \\
        \hline
        \multirow{5}{*}{\tabincell{c}{啼鸣\\(birdsong)}}  &
        \multirow{5}{*}{\tabincell{l}{喊(cry)\\ 禽(bird)}} & Baseline &
        \tabincell{l}{发出哀鸣的声音\\(To make a whining sound.)} \\
        \cline{3-4}
        & & AAM & \tabincell{l}{发出大的声音 \\ (To make a loud noise.)}  \\
        \cline{3-4}
        & & SAAM &\tabincell{l}{鸟类发出的特有的声音 \\ (The unique cry of birds.)}  \\
        \hline
        \multirow{5}{*}{\tabincell{c}{旅馆\\(hotel)}}  & \multirow{5}{*}{\tabincell{l}{场所(place)\\ 旅游(tour)\\ 吃(eat) \\娱乐(recreation) \\住下(reside)}} & Baseline & \tabincell{l}{为人们提供食宿的地方\\(A place to provide residence to people.)} \\
        \cline{3-4}
        & & AAM & \tabincell{l}{为旅行者提供食宿的地方 \\ (A place to provide residence to tourists.)}  \\
        \cline{3-4}
        & & SAAM & \tabincell{l}{给旅行者提供食宿和其他服务的地方 \\ (A place to provide residence and other services to tourists.)}  \\
        \hline
    \end{tabular}
    \caption{Example definitions generated by our models. Baseline represents \citet{Noraset+:2017}. Note that Baseline do not utilize sememes, while the AAM and SAAM models both use sememes.} \label{table:example}
\end{table*}

\begin{table}
    \centering
    \begin{tabular}{lll}
        Model & Valid & Test \\ \hline \hline
        Baseline & 29.57 & 29.71 \\ \hline
        AAM & 32.24 & 32.81  \\
        SAAM & {\bf 37.11} &	{\bf 36.36} \\  
        \ \ \ \ $-$position & 36.79  & 36.05  \\
        \ \ \ \ $-$adaptive & 35.69  & 35.93  \\
        \ \ \ \ $-$sememes & 32.26  & 32.83  \\
    \end{tabular}
    \caption{Ablation study: BLEU scores on the CDM validation set and test set. For the last three rows, we remove position embedding, the adaptive attention layer or sememes information from SAAM model. \label{tab:bleu}}
\end{table}

\begin{figure}[tb]
    \centering
    \includegraphics[width=0.45\textwidth]{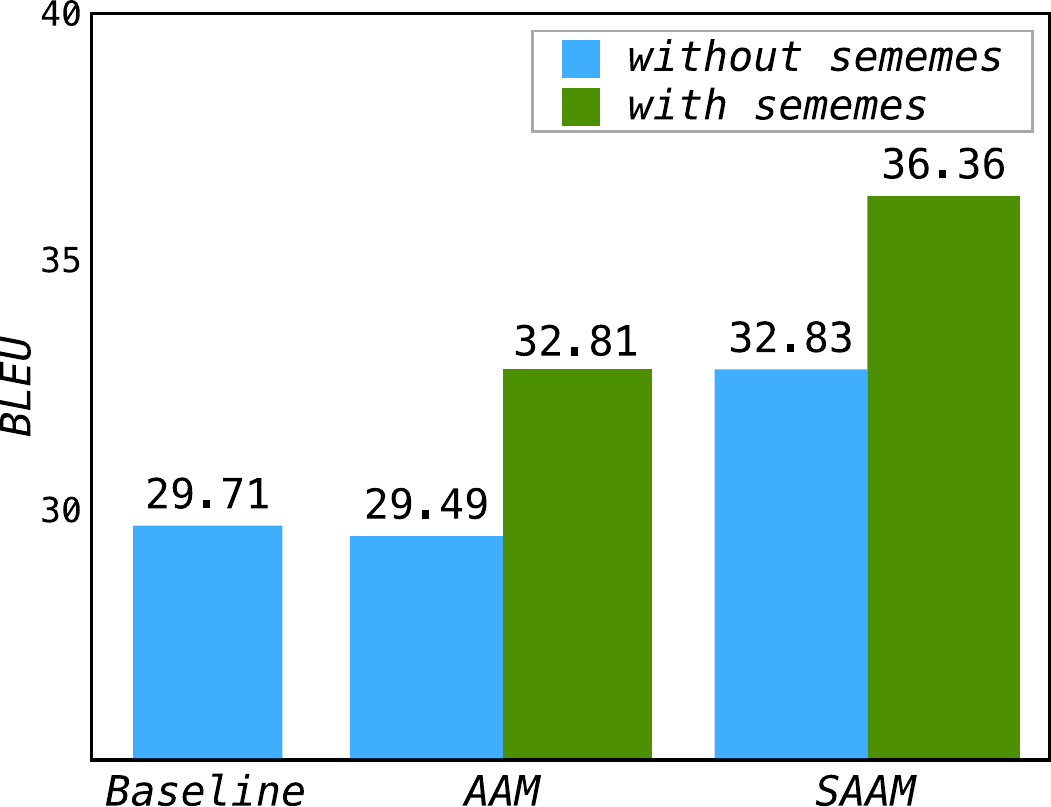}
    \caption{Experimental results of the three models on CDM test set. Since this is the first work to utilize sememes and attention mechanism for definition modeling, the baseline method is non-attention and non-sememes.}\label{fig:sememe}
\end{figure} 

\subsection{Settings}
We show the effectiveness of all models on the CDM dataset. All the embeddings, including word and sememe embedding, are fixed 300 dimensional word embeddings pretrained on the Chinese Gigaword corpus (LDC2011T13). All definitions are segmented with Jiaba Chinese text segmentation tool \footnote{https://github.com/fxsjy/jieba} and we use the resulting unique segments as the decoder vocabulary. 
To evaluate the difference between the generated results and the gold-standard definitions, we compute BLEU score using a script provided by Moses, following \citet{Noraset+:2017}. 
We implement the Baseline and AAM by modifying the code of \citet{Lu+:2017} \footnote{https://github.com/yufengm/Adaptive}, and SAAM with fairseq-py \footnote{https://github.com/pytorch/fairseq}. 

\paragraph{Baseline}
We use two-layer LSTM network as the recurrent component. We set batch size to 128, and the dimension of the hidden state to $300$ for the decoder. Adam optimizer is employed with an initial learning rate of $1\times 10^{-3}$.  Since the morphemes of the word to be defined can benefit definition modeling, \citet{Noraset+:2017} obtain the model with the best performance by adding a trainable embedding from character-level CNN to the fixed word embedding. To obtain the state-of-the-art result as the baseline, we follow \citet{Noraset+:2017} and experiment with the character-level CNN with the same hyperparameters.

\paragraph{AAM}
To be comparable with the baseline, we also use two-layer LSTM network as the recurrent component.We set batch size to 128, and the dimension of the hidden state to $300$ for both the encoder and the decoder. Adam optimizer is employed with an initial learning rate of $1\times 10^{-3}$. 

\paragraph{SAAM}
We have the same hyperparameters as \citet{Vaswani+:2017}, and set these hyperparameters as $(d_{\text{model}}=300, d_{\text{hidden}}=2048, n_{\text{head}}=5, n_{\text{layer}}=6)$. To be comparable with AAM, we use the same batch size as 128. We also employ label smoothing technique \cite{Szegedy+:2016} with a smoothing value of 0.1 during training.

\subsection{Results}
\paragraph{Main Results} We report the experimental results on CDM test set in Figure \ref{fig:sememe}. It shows that both of our proposed models, namely AAM and SAAM, achieve good results and outperform the baseline by a large margin. With sememes, AAM and SAAM can improve over the baseline with +3.1 BLEU and +6.65 BLEU, respectively. 

We also find that sememes are very useful for generating the definition. The incorporation of sememes improves the AAM with +3.32 BLEU and the SAAM with +3.53 BLEU. This can be explained by that sememes help to disambiguate the word sense associated with the target definition.

Among all models, SAAM which incorporates sememes achieves the new state-of-the-art, with a BLEU score of 36.36 on the test set, demonstrating the effectiveness of sememes and the architecture of SAAM. 

Table \ref{table:example} lists some example definitions generated with different models. For each word-sememes pair, the generated three definitions are ordered according to the order: Baseline, AAM and SAAM. For AAM and SAAM, we use the model that incorporates sememes. These examples show that with sememes, the model can generate more accurate and concrete definitions. For example, for the word {``旅馆'' (hotel)}, the baseline model fails to generate definition containing the token  {``旅行者''(tourists)}. However, by incoporating sememes' information, especially the sememe {``旅游'' (tour)}, AAM and SAAM successfully generate {``旅行者''(tourists)}. Manual inspection of others examples also supports our claim.

\paragraph{Ablation Study}
We also conduct an ablation study to evaluate the various choices we made for SAAM. We consider three key components: position embedding, the adaptive attention layer, and the incorporated sememes. As illustrated in table \ref{tab:bleu}, we remove one of these components and report the performance of the resulting model on validation set and test set. We also list the performance of the baseline and AAM for reference. 

It demonstrates that all components benefit the SAAM. Removing position embedding is 0.31 BLEU below the SAAM on the test set. Removing the adaptive attention layer is 0.43 BLEU below the SAAM on the test set. Sememes affects the most. Without incoporating sememes, the performance drops 3.53 BLEU on the test set.

\section{Related Work}
\subsection{Definition Modeling}
Distributed representations of words, or word embeddings \cite{Mikolov+:2013} were widely used in the field of NLP in recent years. Since word embeddings have been shown to capture lexical semantics, \citet{Noraset+:2017} proposed the definition modeling task as a more transparent and direct representation of word embeddings. This work is followed by \citet{Gadetsky+:2018}, who studied the problem of word ambiguities in definition modeling by employing latent variable modeling and soft attention mechanisms. Both works focus on evaluating and interpreting word embeddings. In contrast, we incorporate sememes to generate word sense aware word definition for dictionary compilation.

\subsection{Knowledge Bases}
Recently many knowledge bases (KBs) are established in order to organize human knowledge in structural forms. By providing human experiential knowledge, KBs are playing an increasingly important role as infrastructural facilities of natural language processing.

HowNet \cite{dong2003hownet} is a knowledge base that annotates each concept in Chinese with one or more sememes. HowNet plays an important role in understanding the semantic meanings of concepts in human languages, and has been widely used in word representation learning \cite{Niu+:2017}, word similarity computation \cite{liu2002word} and sentiment analysis \cite{XIANGHUA2013186}. For example, \citet{Niu+:2017} improved word representation learning by utilizing sememes to represent various senses of each word and selecting suitable senses in contexts with an attention mechanism. 

Chinese Concept Dictionary (CCD) is a WordNet-like semantic lexicon \cite{yu2001introduction,Yu2002WSDAC}, where each concept is defined by a set of synonyms (SynSet). CCD has been widely used in many NLP tasks, such as word sense disambiguation \cite{Yu2002WSDAC}.

In this work, we annotate the word with aligned sememes from HowNet and definition from CCD.

\subsection{Self-Attention}
Self-attention is a special case of attention mechanism that relates different positions of a single sequence in order to compute a representation for the sequence. Self-attention has been successfully applied to many tasks recently \cite{Cheng+:2016,Parikh+:2016,Lin+:2017,Vaswani+:2017,Tan+:2018,Kitaev+klein:2018}.

\citet{Vaswani+:2017} introduced the first transduction model based on self-attention by replacing the recurrent layers commonly used in encoder-decoder architectures with multi-head self-attention. The proposed model called Transformer achieved the state-of-the-art performance on neural machine translation with reduced training time. After that, \citet{Tan+:2018} demonstrated that self-attention can improve semantic role labeling by handling structural information and long range dependencies. \citet{Kitaev+klein:2018} further extended self-attention to constituency parsing and showed that the use of self-attention helped to analyze the model by making explicit the manner in which information is propagated between different locations in the sentence.

Self-attention has many good properties. It reduces the computation complexity per layer, allows for more parallelization and reduces the path length between long-range dependencies in the network. In this paper, we use self-attention based architecture in SAAM to learn the relations of word, sememes and definition automatically.

\section{Conclusion}
We introduce the Chinese definition modeling task that generates a definition in Chinese for a given word and sememes of a specific word sense. This task is useful for dictionary compilation. To achieve this, we constructed the CDM dataset with word-sememes-definition triples. We propose two novel methods, AAM and SAAM, to generate word sense aware definition by utilizing sememes. 
In experiments on the CDM dataset we show that our proposed AAM and SAAM outperform the state-of-the-art approach with a large margin. By efficiently incorporating sememes, the SAAM achieves the best performance with improvement over the state-of-the-art method.
\end{CJK*}

\bibliographystyle{acl_natbib}
\bibliography{acl2019}
\end{document}